\documentclass[10pt,twocolumn,letterpaper]{article}

\usepackage{iccv}
\usepackage{times}
\usepackage{epsfig}
\usepackage{graphicx}
\usepackage{amsmath}
\usepackage{amssymb}
\usepackage{url}
\usepackage{bm}
\usepackage{multirow}
\usepackage{booktabs}
\usepackage{geometry}
\usepackage{subcaption}

\newcommand\Tstrut{\rule{0pt}{2.2ex}}         % = `top' strut
\usepackage[pagebackref=true,breaklinks=true,letterpaper=true,colorlinks,bookmarks=false]{hyperref}

\iccvfinalcopy % *** Uncomment this line for the final submission

\def\iccvPaperID{1650} % *** Enter the ICCV Paper ID here

\ificcvfinal\fi

\begin{document}

%%%%%%%%% TITLE
\title{IICNet: A Generic Framework for Reversible Image Conversion}

% \author{Ka Leong Cheng\footnotemark[1]\\
% HKUST\\
% \and
% Yueqi Xie\footnotemark[1]\\
% HKUST\\
% \and
% Qifeng Chen\\
% HKUST
% }

\author{Ka Leong Cheng\footnotemark[1]\ , Yueqi Xie\footnotemark[1]\ , Qifeng Chen\\
The Hong Kong University of Science and Technology\\
{\tt\small \{klchengad, yxieay\}@connect.ust.hk, cqf@ust.hk}
}

% \author{Ka Leong Cheng\footnotemark[1] \qquad Yueqi Xie\footnotemark[1] \qquad Qifeng Chen\\
% The Hong Kong University of Science and Technology\\
% {\tt\small \{klchengad, yxieay\}@connect.ust.hk, cqf@ust.hk}
% }

\twocolumn[{%
\renewcommand\twocolumn[1][]{#1}%
\maketitle
\begin{center}
    \centering
    \includegraphics[scale=0.525]{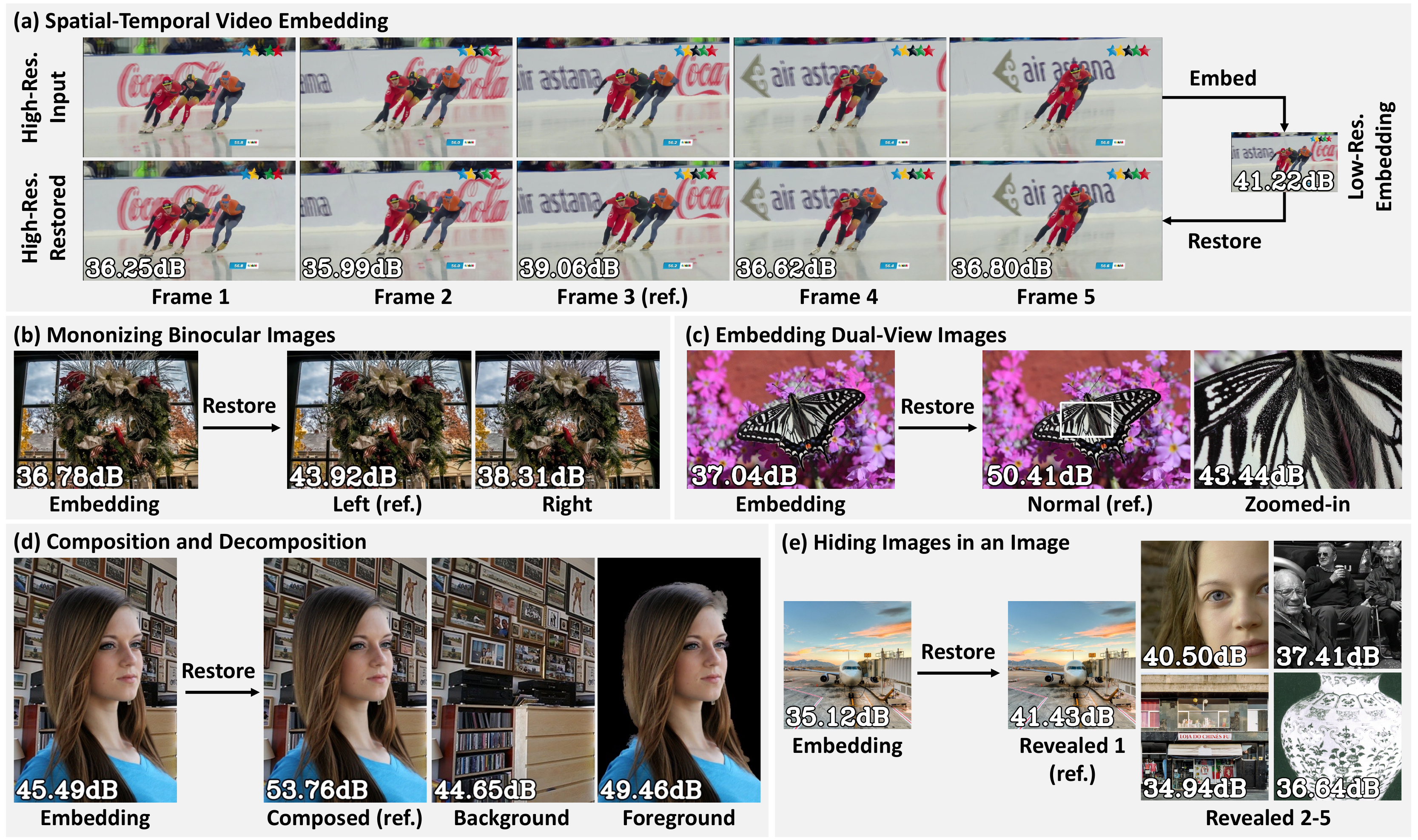}
    \captionof{figure}{(a) Our IICNet can embed a high-resolution sequency into one low-resolution embedding image, which can be used to restore the original content when necessary. (b-e) Our IICNet is the first approach that can generalize among various reversible image conversion (RIC) tasks. We show the whole process of IICNet in (a) but only the restoration process in (b-e).}
    \label{fig:front}
\end{center}
}]
% \maketitle

\renewcommand{\thefootnote}{\fnsymbol{footnote}}
\footnotetext[1]{Joint first authors}
\ificcvfinal\thispagestyle{empty}\fi
%%%%%%%%% ABSTRACT
\begin{abstract}
Reversible image conversion (RIC) aims to build a reversible transformation between specific visual content (e.g., short videos) and an embedding image, where the original content can be restored from the embedding when necessary. This work develops Invertible Image Conversion Net (IICNet) as a generic solution to various RIC tasks due to its strong capacity and task-independent design. Unlike previous encoder-decoder based methods, IICNet maintains a highly invertible structure based on invertible neural networks (INNs) to better preserve the information during conversion. We use a relation module and a channel squeeze layer to improve the INN nonlinearity to extract cross-image relations and the network flexibility, respectively. Experimental results demonstrate that IICNet outperforms the specifically-designed methods on existing RIC tasks and can generalize well to various newly-explored tasks. With our generic IICNet, we no longer need to hand-engineer task-specific embedding networks for rapidly occurring visual content. Our source codes are available at:~\url{https://github.com/felixcheng97/IICNet}.

\end{abstract}
\section{Introduction}
\label{sec:introduction}
Visual media can be classified into different types, including live photos~\cite{live_photo}, binocular images or videos~\cite{hu2020mononizing}, and dual-view images or videos~\cite{dual_view}. Usually, specific devices or platforms are required to view the visual media content. For example, binocular content may only be applicable in 3D devices, so we may need to generate corresponding monocular content to make them compatible with common devices~\cite{hu2020mononizing}. Instead of simply dropping parts of the original content, a better choice is to build a reversible transformation, where the embedding is compatible with common devices, and the original content can be restored when necessary. Also, the single embedding image can help save the storage cost and transmission bandwidth. As a result, many researchers are motivated to study several reversible image conversion (RIC) tasks~\cite{hu2020mononizing,xia2018invertible,zhu2020video} to establish a reversible transformation between visual content and an embedding image. Some examples are shown in Figure~\ref{fig:front}.

RIC tasks are challenging since we often need to embed much richer information implicitly in one single image, which may lead to unavoidable information loss. Previous works~\cite{hu2020mononizing,xia2018invertible,zhu2020video} usually employ an encoder-decoder based framework, which learns the informative bottleneck representation but has limited ability to capture the lost information~\cite{wang2020modeling, xiao2020invertible}. For example, Zhu et al.~\cite{zhu2020video} embed a video preview into a single image and restore the original content with cascaded encoders and decoders, in which they sacrifice the quality of the embedding image to embed more information, but their restored frames are still not highly accurate due to the information loss problem. Hence, one key objective in RIC tasks is to mitigate such information loss. Another concern is that although RIC tasks share the same embedding-restoration procedure for high-quality embedding and restored images, previous methods usually have task-specific designs (e.g., optical flow in~\cite{zhu2020video}), making them challenging to generalize to other types of visual content. Hence, with the rapid growth of media formats plus the increasing interest in the RIC tasks, it is desirable to develop a generic framework for solving all types of RIC tasks.

Considering these aspects, we propose Invertible Image Conversion Net (IICNet) as a generic framework for RIC tasks. To alleviate the information loss problem, we utilize invertible neural networks (INNs)~\cite{dinh2015nice,dinh2017density} as a strictly invertible embedding module. A channel squeeze layer~\cite{xie2021enhanced} is used and integrated into INNs for flexible reduction of dimensions, with only very minor deviations introduced to the invertible architecture. Furthermore, we introduce a relation module to strengthen the limited nonlinear representation capability of INNs~\cite{dinh2015nice} to better capture cross-image relations, in which independent cross-image convolution layers are used, with residual connections for better maintaining a highly reversible structure.

With the strong embedding capacity and the generic module design, IICNet does not rely on any task-specific technique, making it capable of dealing with different content types. We also allow lower-resolution embedding for higher compression rates.

Figure~\ref{fig:front}(a) gives a concrete example for illustration. Given a sequence of video frames, our IICNet can embed the spatial-temporal information of the sequence into one lower-resolution image that is visually similar to the downsampled middle reference frame. There are some promising applications. First, we may embed a short video clip or live photo in one image. Second, we can embed a high-resolution high-FPS video into a low-resolution low-FPS video. In this way, we can allow flexible adoptions for different devices and save storage. Other potential applications are shown in Figure~\ref{fig:front}(b-e), including mononizing binocular images, embedding dual-view images or multi-layer images, and even the general image hiding steganography task.

This paper presents the first generic framework IICNet for different RIC tasks, supported by extensive experiments on five tasks, including two newly-explored tasks: (1) embedding a dual-view image into a single-view one; (2) the reversible conversion between multi-layer images and a single image. Both quantitative and qualitative results show that our method outperforms the existing methods on the studied tasks. Ablation studies are conducted for the network modules and loss functions. More information and demo results are included in the supplementary materials. 

\begin{figure*}[t]
\begin{center}
\includegraphics[scale=0.5]{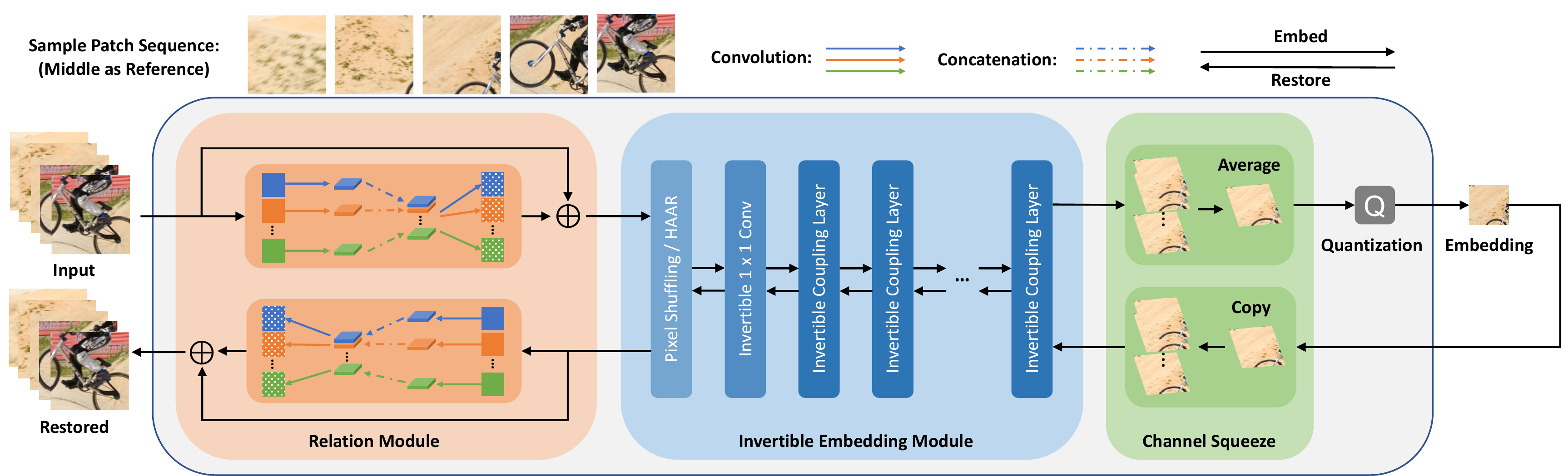}
\end{center}
\caption{Overview of the proposed network. IICNet sequentially contains a relation module, an invertible embedding module (an optional downscaling module plus several coupling layers), a channel squeeze layer, and a quantization layer.}
\label{fig:overview}
\end{figure*}

\section{Related Work}
\label{sec:relatedwork}
\subsection{Reversible Image Conversion}
Our work solves the embedding-and-restoration problem, which belongs to the category of reversible image conversion (RIC). Xia et al.~\cite{xia2018invertible} first propose to encode the original color information into a synthesized grayscale image, from which the color image can be decoded. Recently, Zhu et al.~\cite{zhu2020video} try to embed a sequence of video frames into one image for single image motion expansion. Hu et al.~\cite{hu2020mononizing} further attempt to build an invertible transformation between binocular and monocular views. Although these approaches perform well in their tasks using different technical designs, none of them can generalize to solve all the tasks above due to the task-specific designs. Also, these methods are generally based on an encoder-decoder framework with limited ability to handle the information loss problem.

The reversible property is also explored in steganography, where concealing and recovering the hidden information can be viewed as a reversible task. It aims to hide information within different information carriers like images. Recently, several learning-based methods~\cite{baluja2017hiding, tancik2020stegastamp, wang2019hidinggan, wengrowski2019light, yang2019hiding, zhu2018hidden} leverage the pair of encoder and decoder to hide different kinds of information in images. Still, some works have a limited hiding capacity with some artifacts. In this work, we mainly focus on RIC tasks related to the image carrier only.

\subsection{Invertible Neural Networks}
Invertible neural networks (INNs)~\cite{dinh2015nice,dinh2017density,kingma2018glow} guarantee the invertibility property with a careful mathematical design of network architecture and several invertible operations. In general, the forward process of an INN architecture can learn a bijective mapping between a source domain $x$ to a target domain $y$, with the forward process $f_{\theta}(x) = y$ and the inverse process $f^{-1}_{\theta}(y) = x$. A tractable Jacobian is another great characteristic of INNs to compute the posterior probabilities explicitly for the bijective mapping.

Normalizing Flow based methods~\cite{kobyzev2020normalizing, rezende2015variational} map a complex distribution $x$ with INNs to a latent distribution $z$ (e.g., Gaussian), usually trained by minimizing the unsupervised negative log-likelihood loss. Different from Normalizing Flow based methods, IRN~\cite{xiao2020invertible} maps a high-resolution image to a low-resolution image by utilizing additional latent output variables to capture the lost high-frequency information~\cite{shannon1949communication} with a cross-entropy loss in the image rescaling task. However, the information loss or residual is usually more complex in other general RIC tasks, making the generalization issue a big challenge for IRN. Recent works also investigate the application of INNs on different tasks, such as conditional image super-resolution~\cite{lugmayr2020srflow}, image generation~\cite{ardizzone2019guided, winkler2019learning}, point cloud generation~\cite{pumarola2020cflow}, segmentation tasks~\cite{winkler2019learning}, and image signal processing pipeline~\cite{xing2021invertible}.

\section{Method}
\label{sec:method}
The proposed IICNet for general reversible image conversion (RIC) tasks aims to encode a series of input images into one reversible image (embedding image), which can have either the same or lower resolution. The embedding image can be decoded back to the original inputs with the network backward passing. The key is to use invertible neural networks (INNs) to model such a bijective mapping. An overview of our generic framework is shown in Figure~\ref{fig:overview}.

\subsection{Model Formulation} \label{model-formulation}
Formally, the input of IICNet is a series of $K$ input images $\{\mathbf{i}_k\}_{k=1}^K$ with $\mathbf{i}_k \in \mathbb{R}^{C \times H \times W}$, where $C$, $H$, and $W$ are the image channel number, height, and width, respectively. IICNet can forwardly encode the input images into an embedding image $\mathbf{e}$, which is visually indistinguishable from the reference image $\mathbf{e}_{ref} \in \mathbb{R}^{C_e \times H_e \times W_e}$. Note that the embedding $C_e$, $H_e$, and $W_e$ may be different from $C$, $H$, and $W$. IICNet can then backwardly decode the quantized embedding image $\mathbf{\hat{e}}$ and restore the input images $\{\mathbf{\hat{i}}_k\}_{k=1}^K$. Note that in actual implementations, $K$ input images are stacked along the channel dimension, with input channel size of $N = CK$, denoted as $\mathbf{x}_{1:N} \in \mathbb{R}^{N \times H \times W}$.

\textbf{Relation module.} 
INNs have strong architecture constraints, limiting the nonlinear representation capacity~\cite{dinh2015nice}. Thus, we propose a relation module to add some nonlinear transformation to help capture cross-image relations. To minimize information loss, we add residual connections to greatly preserve the network reversibility.
% this transformation is performed at the same dimensional level with residual connections.

Details of the relation module are shown as the orange part in Figure~\ref{fig:overview}. $K$ parallel convolutional headers independently transform $K$ images into their feature space. The concatenation of the $K$ image features then goes through $K$ independent convolutional tailers plus residual connections to obtain the corresponding images with relational information extracted. The convolutional blocks used here are based on the Dense Block~\cite{huang2017dbnet}. We can express the forward process $f^k_{rel}$ for the $k^{th}$ image $\mathbf{x}_{(kC-C+1):kC}$ as follows:
\begin{equation}
    \mathbf{r}_{(kC-C+1):kC} = f^k_{rel}(\mathbf{x}_{1:N}) + \mathbf{x}_{(kC-C+1):kC}.
\end{equation}
We then obtain $\mathbf{r}_{1:N} \in \mathbb{R}^{N \times H \times W}$. For the inverse process, we apply a symmetric relation module.

\textbf{Invertible downscaling module.} If we optionally activate the invertible downscaling module, IICNet can embed the input images into a lower-resolution embedding image. This module is composed of either a pixel shuffling layer (squeezing operation)~\cite{dinh2017density} or a Haar wavelet transformation layer~\cite{viola2001rapid}, followed by an invertible $1 \times 1$ convolution~\cite{kingma2018glow}. This module offers an invertible operation to halve the resolution of the input images, transforming the size of input tensor from $(N, H, W)$ to $(4N, \frac{1}{2}H, \frac{1}{2}W) = (M, H_e, W_e)$. We describe the forward process of this module $f_{down}$ as:
\begin{equation}
    \mathbf{u}_{1:M} = f_{down}(\mathbf{r}_{1:N}).
\end{equation}
If downscaling is disabled, $f_{down}$ is simply an identical function, yielding $\mathbf{u}_{1:M} = \mathbf{r}_{1:N}$.

\textbf{Coupling layers.} Following the design proposed in~\cite{ardizzone2018analyzing,dinh2015nice,dinh2017density}, we structure a deep INN architecture with several basic inveritble building blocks using two complementary affine coupling layers each. Considering the $l^{th}$ block, the corresponding input tensor $\mathbf{u}_{1:M}$ is split into top parts $\mathbf{u}^l_{t} = \mathbf{u}^l_{1:\tilde{m}}$ and bottom parts $\mathbf{u}^l_{b} = \mathbf{u}^l_{(\tilde{m}+1):M}$ at position $\tilde{m}$. The two corresponding affine transformations are formulated as follows, with element-wise multiplication $\odot$, exponential function $\text{exp}(\cdot)$, and centered sigmoid function $\sigma_c(\cdot) = 2\sigma(\cdot) - 1$:
\begin{equation}
    \mathbf{u}^{l+1}_t = \mathbf{u}^l_t + h_2(\mathbf{u}^l_b),
\end{equation}
\begin{equation}
    \mathbf{u}^{l+1}_b = \mathbf{u}^l_b \odot \text{exp}(\sigma_c(g(\mathbf{u}^{l+1}_t))) + h_1(\mathbf{u}^{l+1}_t).
\end{equation}
Then $\mathbf{u}^{l+1}_t$ and $\mathbf{u}^{l+1}_b$ are concatenated to get $\mathbf{u}^{l+1}_{1:M}$. We can show that the two transformations are invertible:
\begin{equation}
    \mathbf{u}^{l}_b = (\mathbf{u}^{l+1}_b - h_1(\mathbf{u}^{l+1}_t)) \odot \text{exp}(-\sigma_c(g(\mathbf{u}^{l+1}_t))),
\end{equation}
\begin{equation}
    \mathbf{u}^{l}_t = \mathbf{u}^{l+1}_t - h_2(\mathbf{u}^l_b).
\end{equation}
Letting $f_{inn}$ be the forward pass of our INN architecture, the output tensor $\mathbf{v}_{1:M}$ can be formulated as follows:
\begin{equation}
    \mathbf{v}_{1:M} = f_{inn}(\mathbf{u}_{1:M}).
\end{equation}

\begin{figure}[t]
\centering
    \includegraphics[width=1.0\linewidth]{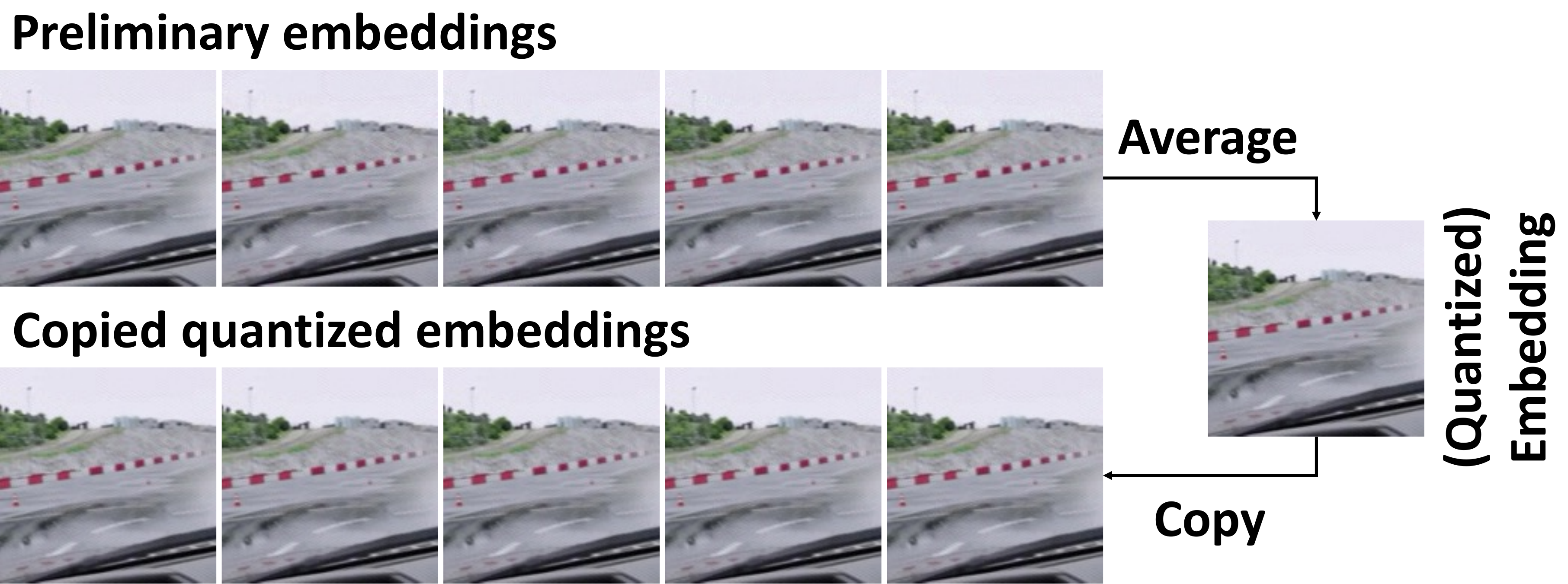}
	\caption{Illustration of the channel squeeze layer.}
	\label{fig:model_cs}
\end{figure}

\textbf{Channel squeeze layer.} Similar to~\cite{xie2021enhanced}, we use a channel squeeze layer but without attention to reduce the channel dimension to obtain the embedding image $\mathbf{e}$. The channel squeeze layer forwardly treats its input tensor $\mathbf{v}_{1:M}$ as a stack of preliminary embedding images $\{\mathbf{e}_k\}_{k=1}^{K_e}$, where $K_e = M / C_e$. The embedding image $\mathbf{e}$ is calculated by averaging the preliminary embedding images:
\begin{equation}
    \mathbf{e} = f_{cs}(\mathbf{v}_{1:M}) = average(\{\mathbf{e}_k\}_{k=1}^{K_e}).
\end{equation}
While for the backward pass, the channel squeeze layer copies the quantized embedding image $\mathbf{\hat{e}}$ multiple times as $\{\mathbf{\hat{e}_k}\}_{k=1}^{K_e}$ and concatenates them along the channel dimension to match the channel size. 

Note that our network is jointly trained as a whole with inherent inverse functions of INNs, and the inverse pass takes the copied (same) quantized embedding images $\{\mathbf{\hat{e}_k}\}_{k=1}^{K_e}$ as input. This implicitly guides the embedding image $\mathbf{e}$ and all the preliminary embedding images $\{\mathbf{e}_k\}_{k=1}^{K_e}$ to look similar to each other. Hence, only minor noise is introduced to the invertibility, and there is no need to pose any explicit constraints on $\{\mathbf{e}_k\}_{k=1}^{K_e}$ during the forward pass. Figure~\ref{fig:model_cs} shows some visual patches of the preliminary embedding images during training, where all the preliminary embedding images are similar to each other. Also, we find that such implicit guidance on the preliminary embedding images helps stabilize the overall training process.

During experiments, we try to pose explicit $L_2$ constraints between $\{\mathbf{e_k}\}_{k=1}^{K_e}$ and $\mathbf{e}$ or to model the information loss for the channel squeeze layer by simple CNNs. But such designs cause worse performance or unstable training. 

\begin{table*}[t]
  \begin{center}
    \begin{tabular}{p{40pt}|p{44pt}p{44pt}p{44pt}p{44pt}p{44pt}p{44pt}p{44pt}p{44pt}}
      \toprule[1pt]
      \multirow{3}{40pt}{\centering\textbf{Step}} & \multicolumn{4}{c}{\textbf{Embedding}} & \multicolumn{4}{c}{\textbf{Restored}} \\
      \ & \multicolumn{2}{c}{Zhu et al.~\cite{zhu2020video}} & \multicolumn{2}{c}{Ours} & \multicolumn{2}{c}{Zhu et al.~\cite{zhu2020video}} & \multicolumn{2}{c}{Ours} \\
      \ & \hfil PSNR & \hfil SSIM & \hfil PSNR & \hfil SSIM & \hfil PSNR & \hfil SSIM & \hfil PSNR & \hfil SSIM \\
      \hline\Tstrut
      \hfil 1 & \hfil 25.277 & \hfil 0.5608 & \hfil \textbf{37.908} & \hfil \textbf{0.9412} & \hfil 34.356 & \hfil 0.9363 & \hfil \textbf{36.698} & \hfil \textbf{0.9519} \\
      \hfil 3 & \hfil 24.561 & \hfil 0.5214 & \hfil \textbf{37.068} & \hfil \textbf{0.9252} & \hfil 33.099 & \hfil 0.9227 & \hfil \textbf{36.302} & \hfil \textbf{0.9490} \\
      \hfil 5 & \hfil 24.246 & \hfil 0.5056 & \hfil \textbf{36.739} & \hfil \textbf{0.9190} & \hfil 32.608 & \hfil 0.9170 & \hfil \textbf{36.074} & \hfil \textbf{0.9475} \\
      \bottomrule[1pt]
    \end{tabular}
  \end{center}
  \caption{Comparison on temporal video embedding test set with embeding range of $9$ and time step of $1$. }
  \label{tab:task1-comparison}
\end{table*}

\textbf{Quantization layer.}
A quantization loss is unavoidable when one saves the embedding image in the common PNG format with only 8 bits per pixel per channel. There are many proposed methods like~\cite{johannes2017end,bengio2013estimating,theis2017lossy} to address this problem. In this paper, we choose to employ the method in~\cite{balle2016end} to add uniform noise during training and do integer rounding during testing to obtain the quantized embedding image $\mathbf{\hat{e}}$. The quantized embedding image further needs to be clamped between $0$ and $255$.

\textbf{Inverse process.} To restore the original input images, we can load the quantized embedding image $\mathbf{\hat{e}}$ and let it sequentially go through the inverse pass of IICNet:
\begin{equation}
    \mathbf{\hat{x}}_{1:N} = (f'_{rel} \circ f^{-1}_{down} \circ f^{-1}_{inn} \circ f'_{cs})(\mathbf{\hat{e}}),
\end{equation}
where $f'_{rel}$, $f^{-1}_{down}$, $f^{-1}_{inn}$, $f'_{cs}$ are the inverse pass functions of the corresponding modules. Then we can obtain the restored images $\{\mathbf{\hat{i}}_k\}_{k=1}^K$.

\subsection{Loss Functions}
As discussed in the channel squeeze layer, we only need to employ loss functions at the two ends: the embedding image and the restored images.

\textbf{Embedding image.} We employ $L_2$ loss to guide the embedding image $\mathbf{e}$ to be visually like the reference image $\mathbf{e}_{ref}$. In the case of downscaling, we use the Bilinear method to downsample the reference image:
\begin{equation}
    \mathcal{L}_{emb} = ||\mathbf{e}_{ref} - \mathbf{e}||^2_2.
\end{equation}

In our experiments, we find that with $L_2$ loss only, the embedding image usually contains many high-frequency patterns. Hence, we further apply one-sided Fourier transform (FT)~\cite{brigham1967fft} on both the embedding image and the reference image to obtain their frequency domain and add a frequency loss $\mathcal{L}_{freq}$ in terms of $L_2$ distance:
\begin{equation}
    \mathcal{L}_{freq} = ||FT(\mathbf{e}_{ref}) - FT(\mathbf{e})||^2_2.
\end{equation}

\textbf{Restored images.} The restored images $\{\mathbf{\hat{i}}_k\}_{k=1}^K$ should match the original ones $\{\mathbf{i}_k\}_{k=1}^K$, so we have another basic restored loss $\mathcal{L}_{res}$ to minimize the average $L_1$ distance among each pair of the restored and original image:
\begin{equation}
    \mathcal{L}_{res} = \frac{1}{K} \sum_{k=1}^{K} ||\mathbf{i}_k - \mathbf{\hat{i}}_k||_1.
\end{equation}

\textbf{Total loss.} To sum up, our proposed IICNet is optimized by minimizing the compact loss $\mathcal{L}_{total}$, with corresponding weight factors $\lambda_1$, $\lambda_2$, $\lambda_3$:
\begin{equation}
    \mathcal{L}_{total} = \lambda_1 \mathcal{L}_{emb} + \lambda_2 \mathcal{L}_{freq} + \lambda_3 \mathcal{L}_{res}.
\end{equation}

\section{Experiments}
\label{sec:experiments}

We first report experiments conducted on the studied RIC tasks in Section~\ref{spatial-temporal-video-embedding} and~\ref{mononizing-binocular-images}, followed by the results of two newly-explored tasks in Section~\ref{dual-view-images} and~\ref{composition-and-decomposition-images}. In Section~\ref{hiding-images-in-an-image}, we try the steganography task to hide several images in one image. The main paper reports multiple-and-single RIC tasks that build a conversion between multiple images and a single image. Our supplements present more results of single-and-single RIC tasks like invertible image rescaling and invertible grayscale. Please also check our supplements for detailed experimental settings.

\subsection{Spatial-Temporal Video Embedding}
\label{spatial-temporal-video-embedding}
The method proposed in~\cite{zhu2020video} aims to embed a sequence of video frames into one embedding image with the same resolution, which can be converted back to the original video sequence. Our proposed IICNet not only performs better but also extends to embed the video frames spatiotemporally into a lower-resolution embedding image. 

\textbf{Dataset and processing.} We use the high-quality DAVIS $2017$ video dataset~\cite{jordi2017davis} in this task. To make our model more robust on different motion levels of video inputs, for each video sample in the train set, we subsample all the possible video subsamples with a time step of $5$ between consecutive frames, where we select the middle frame as the reference image.

\begin{figure}
    \begin{subfigure}{1.0\linewidth}
        \includegraphics[width=0.98\linewidth]{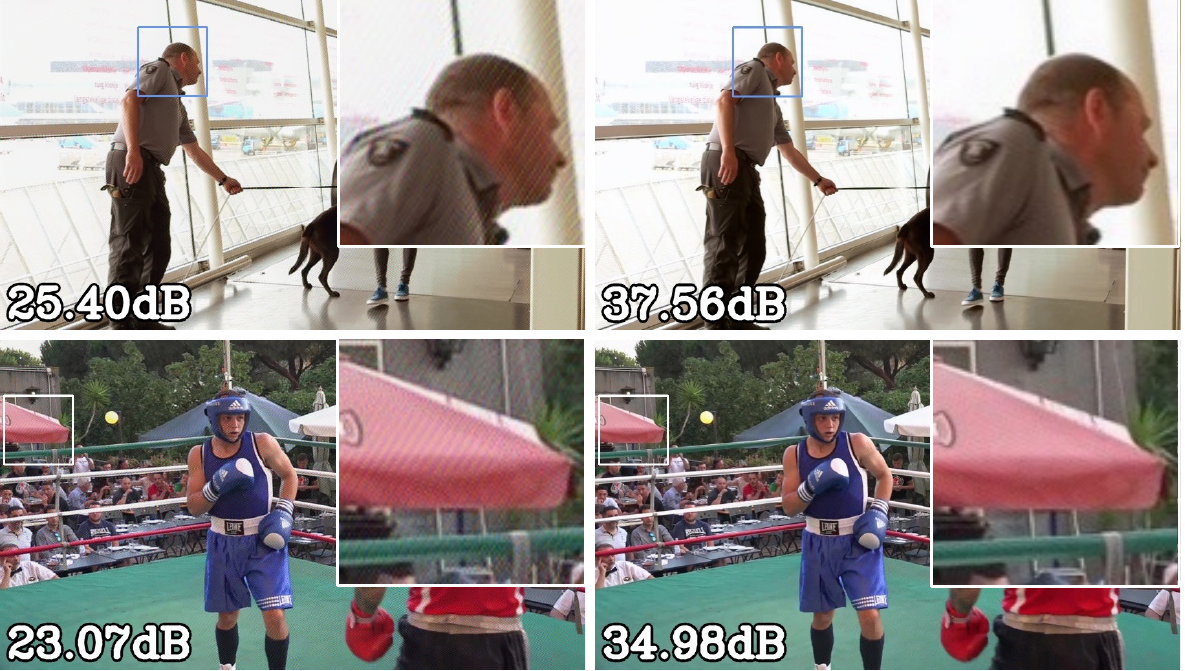}
    \end{subfigure}
    \\[2pt]
    \begin{subfigure}{1.0\linewidth}
        \begin{subfigure}{0.495\linewidth}\hfil Zhu et al.~\cite{zhu2020video}\end{subfigure}
        \begin{subfigure}{0.495\linewidth}\hfil Ours\end{subfigure}
	\end{subfigure}
	\caption{Visual result comparisons on embedding images.}
	\label{fig:task1-embedding}
\end{figure}

\textbf{Result comparison.} Table~\ref{tab:task1-comparison} only reports the comparison results on the test set with embedding range $N = 9$, since the baseline method~\cite{zhu2020video} only provides the pre-trained $N = 9$ model. We study the performance at different time step levels of $1$, $3$, $5$ to test the capacity of the models in handling small and large motions. Statistics show that our method significantly outperforms the baseline method at all time step levels by large margins. Without dependence on optical flow, our method has less performance drop as the time step grows. We also offer grayscale PSNR and SSIM comparisons in our supplements for reference.

\begin{figure*}
    \begin{subfigure}{0.02\linewidth}
        \parbox[][1.9cm][c]{\linewidth}{\centering\raisebox{0in}{\rotatebox{90}{Zhu et al.~\cite{zhu2020video}}}} \\
        \parbox[][1.9cm][c]{\linewidth}{\centering\raisebox{0in}{\rotatebox{90}{Ours}}} \\
        \parbox[][1.9cm][c]{\linewidth}{\centering\raisebox{0in}{\rotatebox{90}{GT}}}
    \end{subfigure}
    \begin{subfigure}{0.98\linewidth}
        \begin{subfigure}{1\linewidth}
            \includegraphics[width=1\linewidth]{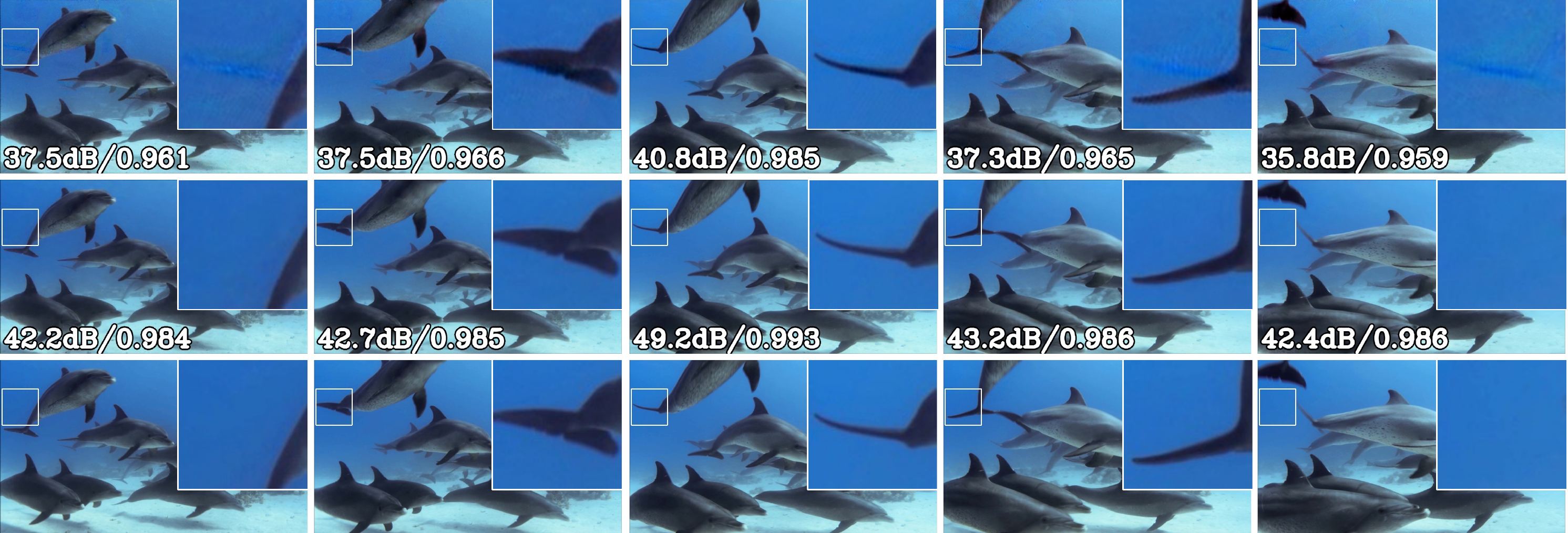}
        \end{subfigure}
    \end{subfigure}
    \\[2pt]
    \begin{subfigure}{0.02\linewidth}\hfil \end{subfigure}
    \begin{subfigure}{0.98\linewidth}
        \begin{subfigure}{0.195\linewidth}\hfil Frame 1\end{subfigure}
        \begin{subfigure}{0.195\linewidth}\hfil Frame 3\end{subfigure}
        \begin{subfigure}{0.195\linewidth}\hfil Frame 5 (ref.)\end{subfigure}
        \begin{subfigure}{0.195\linewidth}\hfil Frame 7\end{subfigure}
        \begin{subfigure}{0.195\linewidth}\hfil Frame 9\end{subfigure}
    \end{subfigure}
    \caption{Visual result comparisons on restored frames.}
	\label{fig:task1-restore}
\end{figure*}

Figure~\ref{fig:task1-embedding} and Figure~\ref{fig:task1-restore} show the visualization results of the embedding images and the restored frames, respectively. Evident artifacts are found in baseline results, especially for the embedding image. In contrast, our embedding and restored images contain very few artifacts, demonstrating the effectiveness of the employed INN architecture in RIC tasks.

\begin{table}
  \begin{center}
    \begin{tabular}{p{29pt}|p{39pt}p{35pt}p{39pt}p{35pt}}
      \toprule[1pt]
      % \multirow{2}{29pt}{\centering\textbf{Range}} 
      \hfil \textbf{Range}& \multicolumn{2}{c}{\textbf{Embedding}} & \multicolumn{2}{c}{\textbf{Restored}} \\ 
      \hfil \textbf{(Res.)} & \hfil PSNR & \hfil SSIM & \hfil PSNR & \hfil SSIM \\
      \hline\Tstrut
      \hfil 5 & \hfil 38.900 & \hfil 0.9522 & \hfil 41.729 & \hfil 0.9807 \\
      \hfil 7 & \hfil 38.157 & \hfil 0.9437 & \hfil 38.785 & \hfil 0.9660 \\
      \hfil 9 & \hfil 37.908 & \hfil 0.9412 & \hfil 36.698 & \hfil 0.9519 \\
      \hline\Tstrut
      \hfil 3 ($\mathbf{\times}$2) & \hfil 37.585 & \hfil 0.9584 & \hfil 36.914 & \hfil 0.9540 \\
      \hfil 5 ($\mathbf{\times}$2) & \hfil 36.692 & \hfil 0.9477 & \hfil 33.977 & \hfil 0.9205 \\
      \bottomrule[1pt]
    \end{tabular}
  \end{center}
  \caption{Results on spatial-temporal video embedding test set with different embedding ranges and resolutions.}
  \label{tab:task1-range}
\end{table}

\textbf{Embedding ranges and resolutions.} To investigate the embedding capacity of our method, we conduct experiments using different embedding ranges ($5, 7, 9$ input images) in Table~\ref{tab:task1-range}. Similarly, we subsample the training videos with a time step of $5$ and test at a time step level of $1$. Intuitively, more input images indicate more challenges because there is usually more motion information to embed into the embedding image. Table~\ref{tab:task1-range} further shows the experimental results of our method to embed the input video sequence spatially and temporally into a lower-resolution embedding image. To the best of our knowledge, no previous work tries to do the spatial-temporal embedding task. We report the results of embedding $N = 3, 5$ into a $2$ times lower-resolution image, conducted with a time step of $5$ and $1$ for training and testing, respectively. We can see that even we compress the input frames up to $20$ times smaller; still, the model can have a good preview image and restored frames.

\begin{figure*}[ht!]
    \begin{subfigure}{0.02\linewidth}
        \parbox[][3.7cm][c]{\linewidth}{\centering\raisebox{0in}{\rotatebox{90}{L. Bino-view}}} \\
        \parbox[][3.7cm][c]{\linewidth}{\centering\raisebox{0in}{\rotatebox{90}{R. Bino-view}}}
    \end{subfigure}
    \begin{subfigure}{0.98\linewidth}
        \includegraphics[width=1\linewidth]{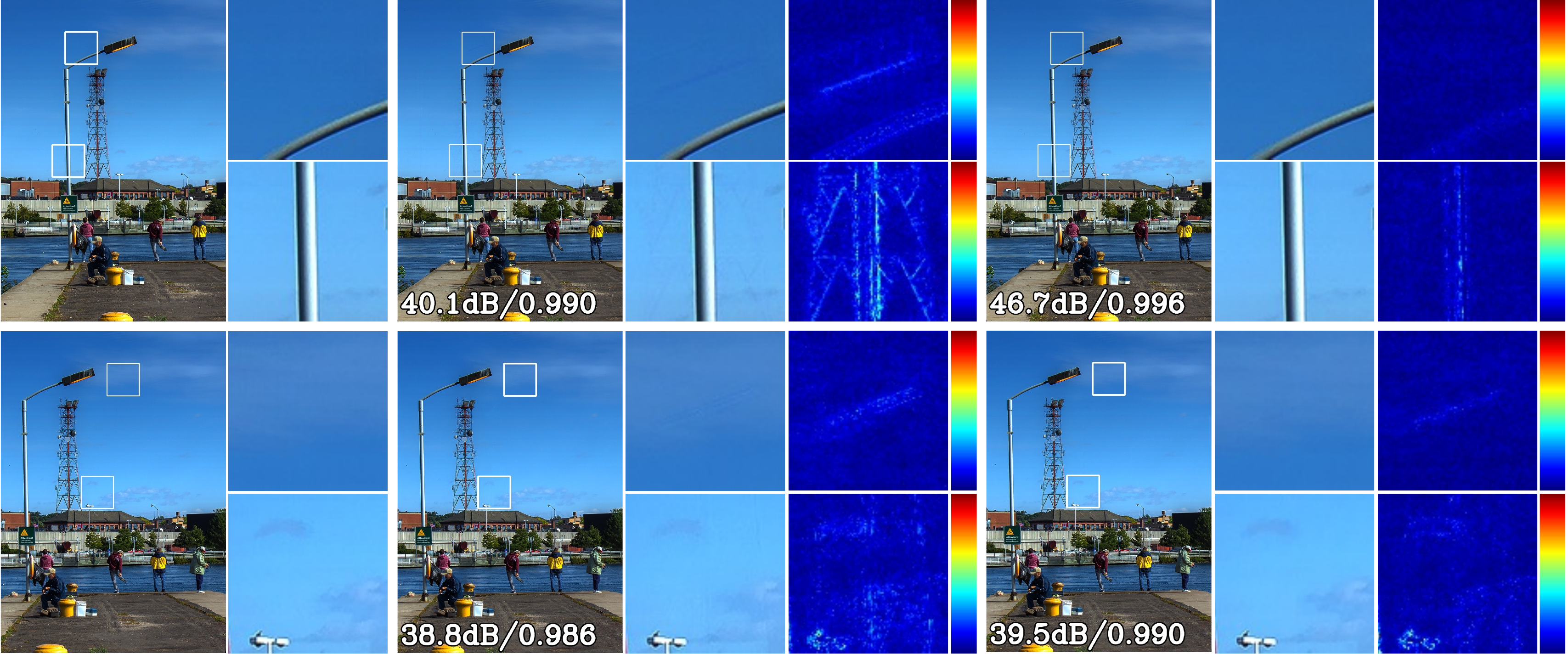}%
    \end{subfigure}
    \\[2pt]
    \begin{subfigure}{0.02\linewidth}\hfil \end{subfigure}
    \begin{subfigure}{0.98\linewidth}
        \begin{subfigure}{0.256\linewidth}\hfil GT\end{subfigure}
        \begin{subfigure}{0.372\linewidth}\hfil Hu et al.~\cite{hu2020mononizing}\end{subfigure}
        \begin{subfigure}{0.372\linewidth}\hfil Ours\end{subfigure}
    \end{subfigure}
	\caption{Visual comparison results of the restored Binocular Images. We show the zoomed-in patches with the corresponding error map aside. Note that we amplified the error maps by 10 times for better visualization.}
	\label{fig:task2-comparison}
\end{figure*}

\subsection{Mononizing Binocular Images}
\label{mononizing-binocular-images}
We also experiment on another studied mononizing binocular images task~\cite{hu2020mononizing}, which aims to convert binocular images or videos into monocular ones with the stereo information implicitly encoded. In this way, monocular devices can cope with stereoscopic data, and the original stereo content can be restored when necessary. We demonstrate that our framework outperforms state-of-the-art methods. 

Same as~\cite{hu2020mononizing}, we train on the Flickr1024 dataset~\cite{wang2019flickr1024} with the official train and test splits. Quantitative results are shown in Table~\ref{tab:task2-comparison}. We achieve the best performance, especially for the restored images, with an improvement of $6.6$dB for the left views and $1.1$dB for the right views. Although Mono3D already achieves good performance, we can still see some structural artifacts like the street lamp and the electric tower in the zoomed-in restored patches, as shown in Figure~\ref{fig:task2-comparison}. In contrast, our method can restore nearly artifact-free binocular views. Although we only train our network on images, results show strong temporal consistency when we apply our model to videos in a per-frame manner. Some demos are in the supplementary video.

% \begin{table}[t!]
%   \begin{center}
%     \begin{tabular}{p{20pt}|p{56pt}|p{56pt}|p{56pt}}
%       \toprule[1pt]
%       \multirow{2}{20pt}{\ } & \multicolumn{1}{c|}{\textbf{Mono-view}} & \multicolumn{1}{c|}{\textbf{L. Bino-view}} & \multicolumn{1}{c}{\textbf{R. Bino-view}} \\ 
%       \ & \hfil PSNR \text{  } SSIM& \hfil PSNR \text{  } SSIM & \hfil PSNR \text{  } SSIM \\
%       \hline\Tstrut
%       \hfil \cite{baluja2017hiding} & \hfil 26.1\ \ \ \ \ \ 0.81 & \hfil -\ \ \ \ \ \ \ \ \ \ \ \ - & \hfil 27.9\ \ \ \ \ \ 0.88 \\
%       \hfil \cite{xia2018invertible} & \hfil 28.0\ \ \ \ \ \ 0.89 & \hfil 28.7\ \ \ \ \ \ 0.92 & \hfil 30.7\ \ \ \ \ \ 0.92 \\
%       \hfil \cite{hu2020mononizing}  & \hfil \textbf{37.8}\ \ \ \ \ \ \textbf{0.97} & \hfil 38.3\ \ \ \ \ \ 0.99 & \hfil 37.3\ \ \ \ \ \ 0.98 \\
%       \hfil Ours & \hfil 37.5\ \ \ \ \ \ 0.95 & \hfil \textbf{44.9}\ \ \ \ \ \ \textbf{0.99} & \hfil \textbf{38.4}\ \ \ \ \ \ \textbf{0.98} \\
%       \bottomrule[1pt]
%     \end{tabular}
%   \end{center}
%   \caption{Results on mononizing binocular images test set.}
%   \label{tab:task2-comparison}
% \end{table}

\begin{table}[t!]
  \begin{center}
    \begin{tabular}{@{}p{58pt}@{}|@{}p{30pt}@{}p{30pt}@{}|@{}p{30pt}@{}p{30pt}@{}|@{}p{30pt}@{}p{30pt}@{}}
      \toprule[1pt]
      \multirow{2}{58pt}{\ } & \multicolumn{2}{@{}c@{}|}{\textbf{Mono-view}} & \multicolumn{2}{@{}c@{}|}{\textbf{L. Bino-view}} & \multicolumn{2}{@{}c@{}}{\textbf{R. Bino-view}} \\ 
      \ & \hfil PSNR & \hfil SSIM & \hfil PSNR & \hfil SSIM & \hfil PSNR & \hfil SSIM \\
      \hline\Tstrut
      \hfil Baluja~\cite{baluja2017hiding}  & \hfil 26.1 & \hfil 0.81 & \hfil -    & \hfil -    & \hfil 27.9 & \hfil 0.88 \\
      \hfil Xia et al.~\cite{xia2018invertible} & \hfil 28.0 & \hfil 0.89 & \hfil 28.7 & \hfil 0.92 & \hfil 30.7 & \hfil 0.92 \\
      \hfil Hu et al.~\cite{hu2020mononizing}  & \hfil \textbf{37.8} & \hfil \textbf{0.97} & \hfil 38.3 & \hfil 0.99 & \hfil 37.3 & \hfil 0.98 \\
      \hfil Ours                     & \hfil 37.5 & \hfil 0.95 & \hfil \textbf{44.9} & \hfil \textbf{0.99} & \hfil \textbf{38.4} & \hfil \textbf{0.98} \\
      \bottomrule[1pt]
    \end{tabular}
  \end{center}
  \caption{Results on mononizing binocular images test set.}
  \label{tab:task2-comparison}
\end{table}

\subsection{Embedding Dual-View Images}
\label{dual-view-images}
Dual-view camera mode is an advanced technology in the field of smartphone cameras, which is first available on HUAWEI P30 Pro~\cite{dual_view}. Users can record split-screen images or videos with the primary camera capturing normal-view images or videos on the left and the zoom lens capturing zoomed-view ($\times 4$) images or videos on the right. 

Similarly, not all devices support dual-view images. Our method can serve as a backward-compatible solution to embed the dual-view images into one normal-view image. We train and test our method using pairs of zoomed-view ($\times 2$,$\times 4$,$\times 8$) and normal-view images generated from the DIV2K dataset~\cite{agustsson2017ntire}, with the normal view images as reference. Some setting details are in the supplements.

\begin{figure}
    \begin{subfigure}{0.035\linewidth}
        \parbox[][][c]{\linewidth}{\centering\raisebox{0in}{\rotatebox{90}{GT}}}
    \end{subfigure}
    \begin{subfigure}{0.965\linewidth}
        \begin{subfigure}{0.33\linewidth}
			\includegraphics[width=0.98\linewidth]{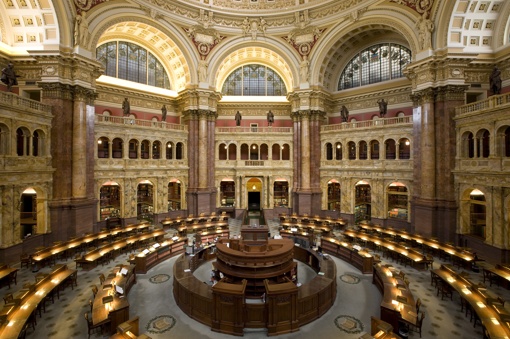}
    	\end{subfigure}%
        \begin{subfigure}{0.33\linewidth}
			\includegraphics[width=0.98\linewidth]{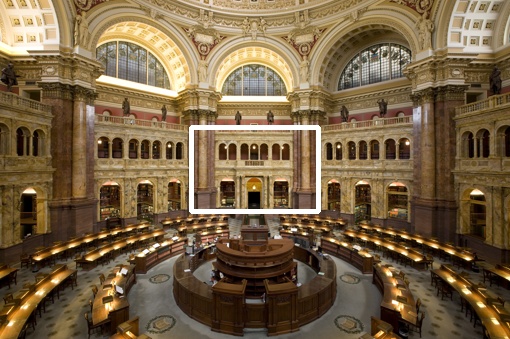}
    	\end{subfigure}%
        \begin{subfigure}{0.33\linewidth}
			\includegraphics[width=0.98\linewidth]{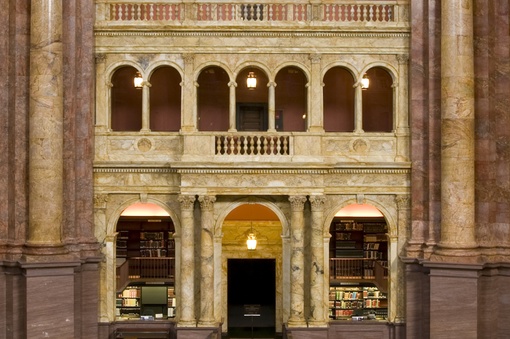}
    	\end{subfigure}%
    \end{subfigure}
    \\
    \begin{subfigure}{0.035\linewidth}
        \parbox[][][c]{\linewidth}{\centering\raisebox{0in}{\rotatebox{90}{Ours}}}
    \end{subfigure}
    \begin{subfigure}{0.965\linewidth}
        \begin{subfigure}{0.33\linewidth}
			\includegraphics[width=0.98\linewidth]{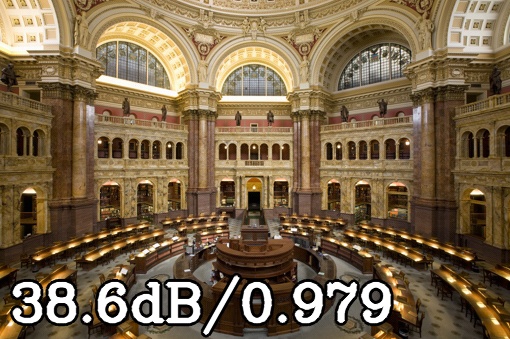}
    	\end{subfigure}%
        \begin{subfigure}{0.33\linewidth}
			\includegraphics[width=0.98\linewidth]{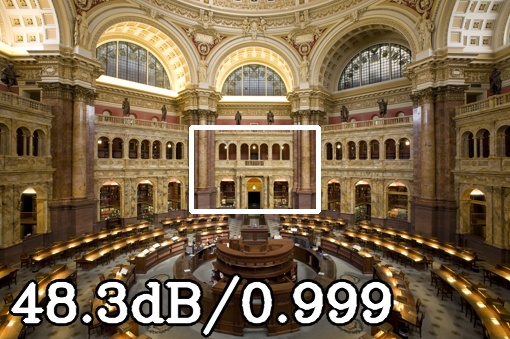}
    	\end{subfigure}%
        \begin{subfigure}{0.33\linewidth}
			\includegraphics[width=0.98\linewidth]{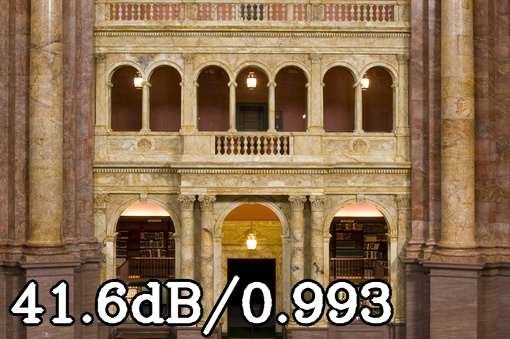}
    	\end{subfigure}%
    \end{subfigure}
    \\[2pt]
    \begin{subfigure}{0.035\linewidth}\hfil \end{subfigure}
    \begin{subfigure}{0.965\linewidth}
        \begin{subfigure}{0.32\linewidth}\hfil Embedding\end{subfigure}
        \begin{subfigure}{0.32\linewidth}\hfil Normal-view\end{subfigure}
        \begin{subfigure}{0.32\linewidth}\hfil Zoomed-view\end{subfigure}
    \end{subfigure}
	\caption{A sample result of embedding dual-view images.}
	\label{fig:application1-visual}
\end{figure}

Quantitative results in Table~\ref{tab:application1-psnr} show that our method achieves great performance to embed dual-view images in terms of both PSNR and SSIM. We also show some visual results in Figure~\ref{fig:application1-visual}, where we can see that both the embedding and restored images are nearly perfect. 

\begin{table}
  \begin{center}
    \begin{tabular}{p{35pt}|p{50pt}p{50pt}p{50pt}}
      \toprule[1pt]
      \hfil \textbf{Modes} & \hfil \textbf{Embedding} & \hfil \textbf{Normal} & \hfil \textbf{Zoomed} \\
      \hline\Tstrut
      \hfil $\times$ 2 & \hfil 38.248 & \hfil 50.171 & \hfil 43.461  \\
      \hfil $\times$ 4 & \hfil 38.438 & \hfil 49.116 & \hfil 43.662  \\
      \hfil $\times$ 8 & \hfil 38.356 & \hfil 48.854 & \hfil 43.578  \\
      \bottomrule[1pt]
    \end{tabular}
  \end{center}
  \caption{PSNR on embedding dual-view images.}
  \label{tab:application1-psnr}
\end{table}

\subsection{Composition and Decomposition}
\label{composition-and-decomposition-images}
Photoshop~\cite{photoshop} is a popular image editing software, where users can use multiple layers to perform tasks such as compositing multiple images into one. Usually, the composition process is not reversible, so we cannot recover the sheltered part of the background in the composed image. However, with our method, we can allow the ``composed image" to embed all the layer images. In this way, although we only store and transmit one ``composed image" as before, users can also get the original layers for further usage.

\begin{table}
  \begin{center}
    \begin{tabular}{p{27pt}|p{38pt}p{38pt}p{38pt}p{38pt}}
      \toprule[1pt]
      \ & \hfil \textbf{Embed.} & \hfil \textbf{Comp.} &  \hfil \textbf{Fg.} & \hfil \textbf{Bg.} \\
      \hline\Tstrut
      \hfil Adobe & \hfil 45.305 & \hfil 52.709 & \hfil 44.586 & \hfil 44.921 \\
      \hfil Real & \hfil 47.350 & \hfil 60.234 & \hfil - & \hfil 43.718 \\
      \bottomrule[1pt]
    \end{tabular}
  \end{center}
  \caption{PSNR on composition and decomposition.}
  \label{tab:application2-psnr}
\end{table}

\begin{figure}
    \begin{subfigure}{0.035\linewidth}
        \parbox[][][c]{\linewidth}{\centering\raisebox{0in}{\rotatebox{90}{GT}}}
    \end{subfigure}
    \begin{subfigure}{0.965\linewidth}
        \begin{subfigure}{0.25\linewidth}
			\includegraphics[width=0.985\linewidth]{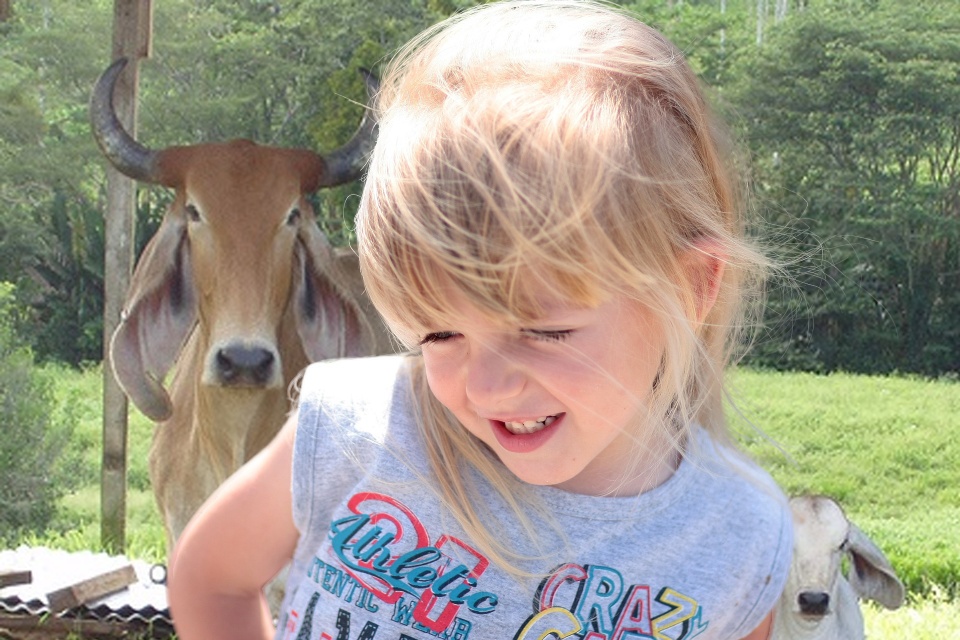}
    	\end{subfigure}%
        \begin{subfigure}{0.25\linewidth}
			\includegraphics[width=0.985\linewidth]{./results/adobe/gt_compose.jpeg}
    	\end{subfigure}%
        \begin{subfigure}{0.25\linewidth}
			\includegraphics[width=0.985\linewidth]{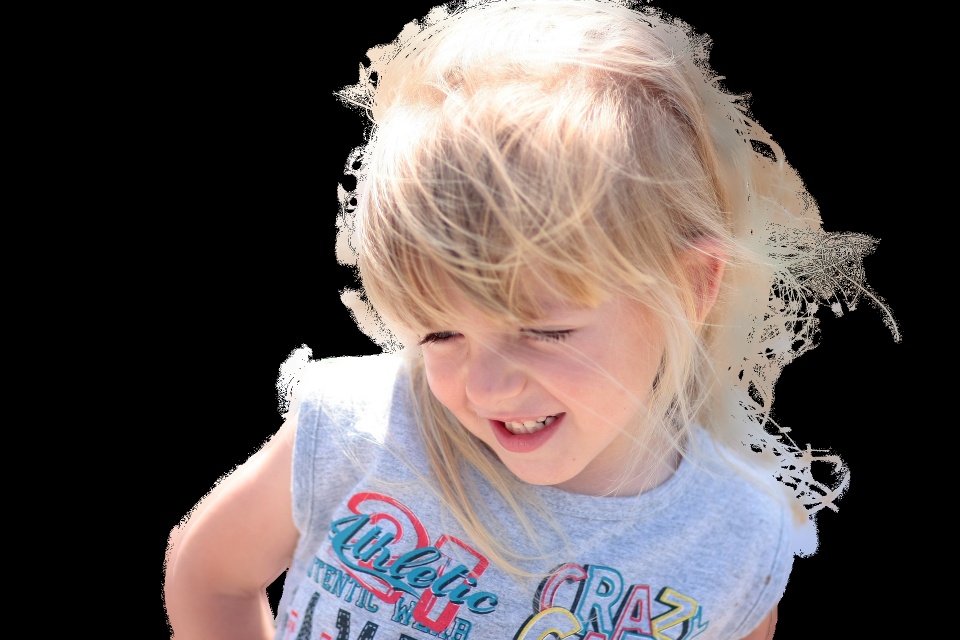}
    	\end{subfigure}%
        \begin{subfigure}{0.25\linewidth}
			\includegraphics[width=0.985\linewidth]{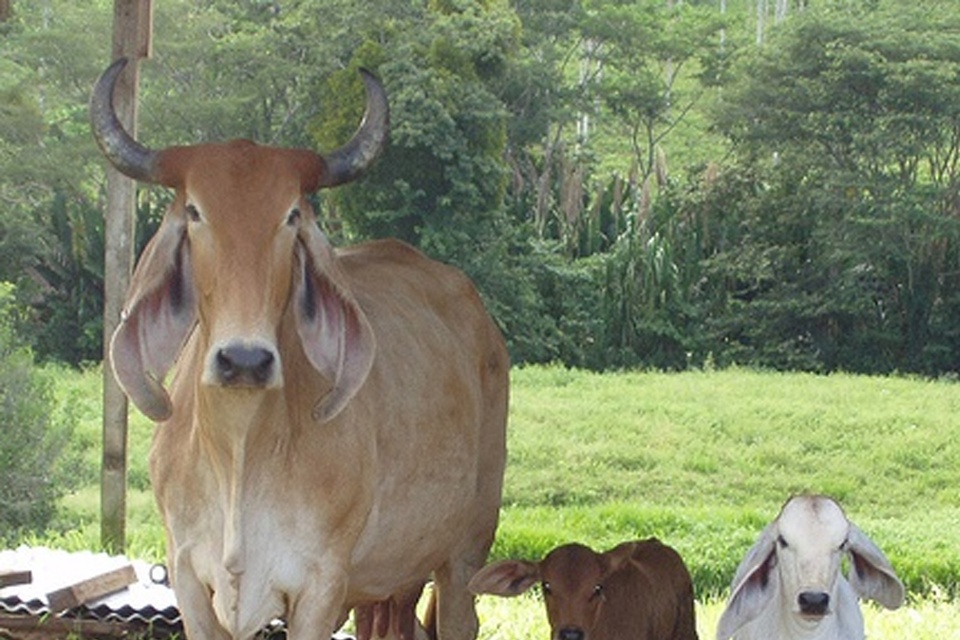}
		\end{subfigure}
	\end{subfigure}
	\\
    \begin{subfigure}{0.035\linewidth}
        \parbox[][][c]{\linewidth}{\centering\raisebox{0in}{\rotatebox{90}{Ours}}}
    \end{subfigure}
    \begin{subfigure}{0.965\linewidth}
        \begin{subfigure}{0.25\linewidth}
			\includegraphics[width=0.985\linewidth]{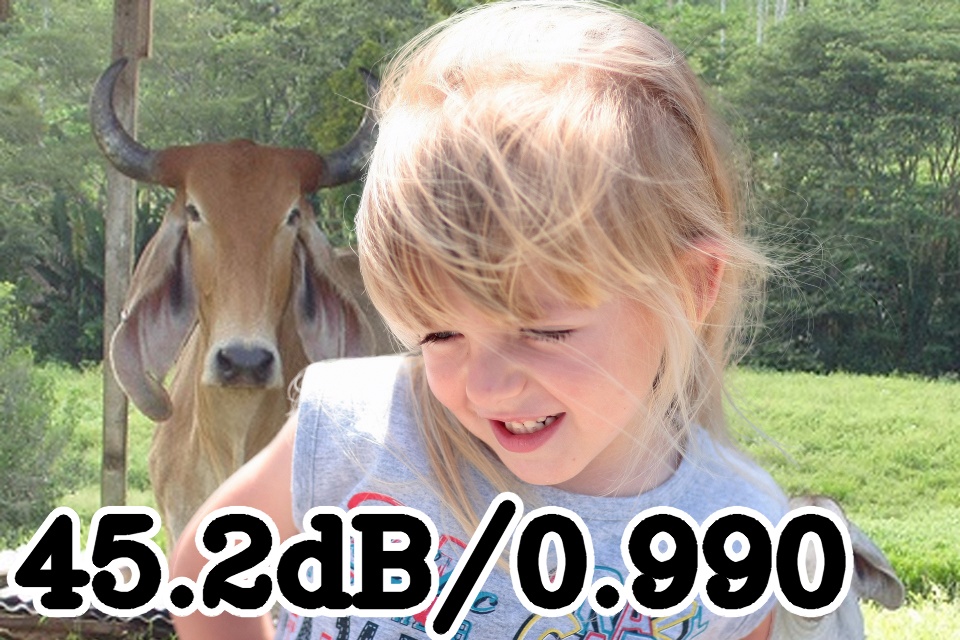}
    	\end{subfigure}%
    	\begin{subfigure}{0.25\linewidth}
			\includegraphics[width=0.985\linewidth]{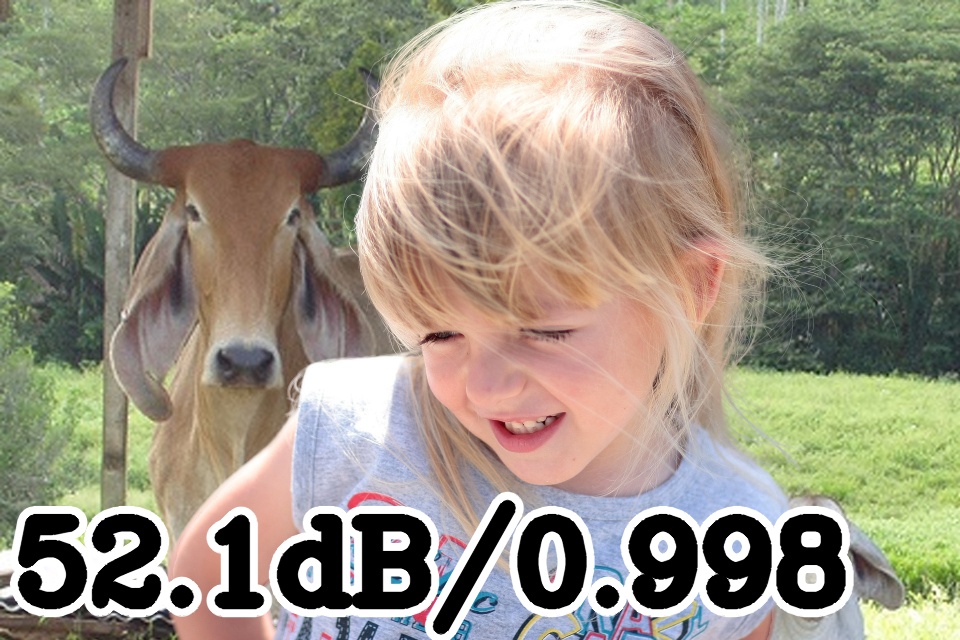}
    	\end{subfigure}%
        \begin{subfigure}{0.25\linewidth}
			\includegraphics[width=0.985\linewidth]{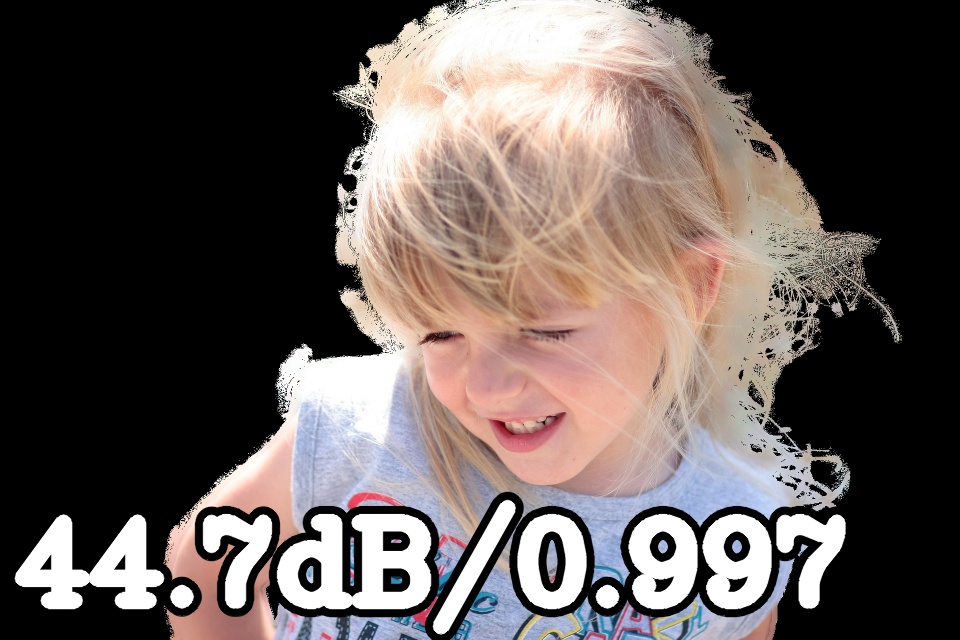}
    	\end{subfigure}%
        \begin{subfigure}{0.25\linewidth}
			\includegraphics[width=0.985\linewidth]{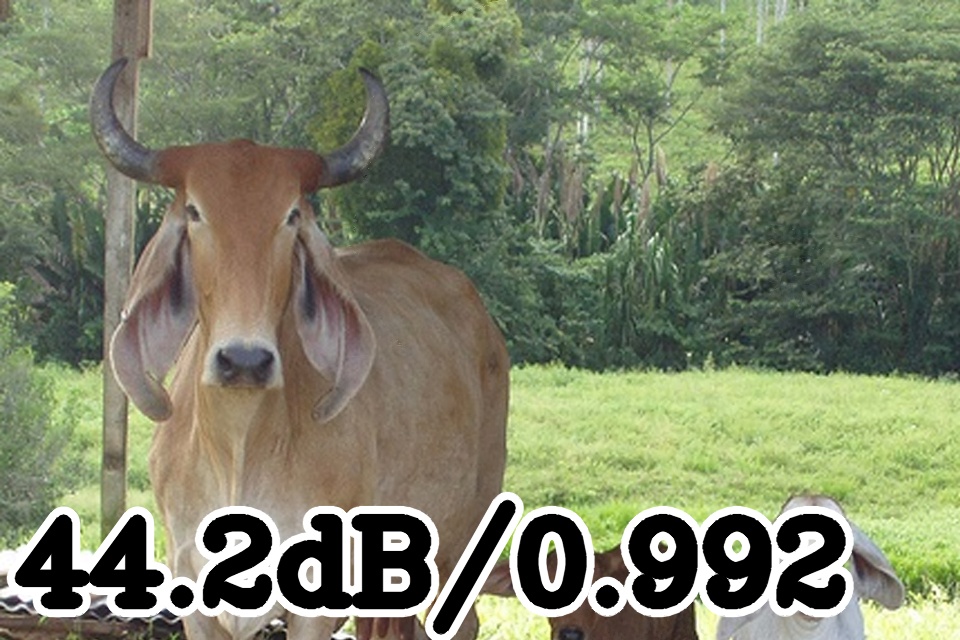}
    	\end{subfigure}%
    \end{subfigure}
    \\[2pt]
    \begin{subfigure}{0.035\linewidth}\hfil \end{subfigure}
    \begin{subfigure}{0.965\linewidth}
        \begin{subfigure}{0.24\linewidth}\hfil Embedding\end{subfigure}
        \begin{subfigure}{0.24\linewidth}\hfil Composed\end{subfigure}
        \begin{subfigure}{0.24\linewidth}\hfil Foreground\end{subfigure}
        \begin{subfigure}{0.24\linewidth}\hfil Background\end{subfigure}
    \end{subfigure}
	\caption{A sample result of composition and decomposition.}
	\label{fig:application2-visual}
\end{figure}

\begin{table*}[t!]
  \begin{center}
    \begin{tabular}{@{}p{60pt}@{}|p{36pt}p{36pt}p{36pt}|p{36pt}p{36pt}p{36pt}|p{36pt}p{36pt}p{36pt}}
      \toprule[1pt]
      \multirow{2}{60pt}{\centering\textbf{Methods}} & \multicolumn{3}{c}{\textbf{Video Embedding}} & \multicolumn{3}{|c}{\textbf{Mononizing Binocular Images}} & \multicolumn{3}{|c}{\textbf{Hiding Images in an Image}} \\ 
      \ & \hfil Embed. & \hfil Restore & \hfil \#Param. & \hfil Embed. & \hfil Restore & \hfil \#Param. & \hfil Embed. & \hfil Restore & \hfil \#Param.\\
      \hline\Tstrut
      \hfil AE~\cite{xia2018invertible}    & \hfil 37.925 & \hfil 37.242 & \hfil 7.43M & \hfil 35.387 & \hfil 38.239 & \hfil 4.55M & \hfil 34.248 & \hfil 31.721 & \hfil 7.43M \\
      \hfil INNs~\cite{xiao2020invertible} & \hfil 34.029 & \hfil 38.452 & \hfil 6.57M & \hfil 34.465 & \hfil 38.171 & \hfil 4.49M & \hfil 29.953 & \hfil 33.843 & \hfil 6.57M \\
      \hfil Ours w/o rel.                  & \hfil 38.752 & \hfil 41.159 & \hfil 6.57M & \hfil 36.684 & \hfil 39.667 & \hfil 4.49M & \hfil 35.533 & \hfil 36.698 & \hfil 6.57M \\
      \hfil Ours w/o freq.                 & \hfil 32.914 & \hfil \textbf{42.353} & \hfil 6.81M & \hfil 31.469 & \hfil 41.161 & \hfil 4.40M & \hfil 28.780 & \hfil 37.623 & \hfil 6.81M \\
      \hfil Ours                           & \hfil \textbf{38.900} & \hfil 41.729 & \hfil 6.81M & \hfil \textbf{37.540} & \hfil \textbf{41.649} & \hfil 4.40M & \hfil \textbf{35.641} & \hfil \textbf{37.935} & \hfil 6.81M \\
      \bottomrule[1pt]
    \end{tabular}
  \end{center}
  \caption{Ablation studies on three representative tasks.}
  \label{tab:ablation}
\end{table*}

\begin{table}[t!]
  \begin{center}
    \begin{tabular}{p{41pt}|p{33pt}p{33pt}p{33pt}p{33pt}}
      \toprule[1pt]
      \multirow{2}{41pt}{\centering\textbf{\#Embed.}} & \multicolumn{2}{c}{\textbf{Embedding}} & \multicolumn{2}{c}{\textbf{Restored}} \\ 
      \ & \hfil PSNR & \hfil SSIM & \hfil PSNR & \hfil SSIM \\
      \hline\Tstrut
      \hfil 2 & \hfil 38.586 & \hfil 0.9403 & \hfil 48.599 & \hfil 0.9945 \\
      \hfil 3 & \hfil 37.038 & \hfil 0.9166 & \hfil 42.884 & \hfil 0.9852 \\
      \hfil 4 & \hfil 36.184 & \hfil 0.9041 & \hfil 39.883 & \hfil 0.9745 \\
      \hfil 5 & \hfil 35.641 & \hfil 0.8913 & \hfil 37.935 & \hfil 0.9638 \\
      \bottomrule[1pt]
    \end{tabular}
  \end{center}
  \caption{PSNR on hiding images in an image.}
  \label{tab:application3-psnr-ssim}
\end{table}

Since there is no publicly available dataset for composition and decomposition, we instead train and test our method on two matting datasets: the Adobe Deep Matting dataset~\cite{xu2017deep} and the Real Matting dataset~\cite{sengupta2020background}. Note that the Real Matting dataset does not have ground truth for the foreground. Detailed settings are available in supplements.

Table~\ref{tab:application2-psnr} shows the quantitative performance of our method on the two datasets. We also include some visual results in Figure~\ref{fig:application2-visual}. We can see that our method performs well and is verified as applicable to the task of composing and decomposing images.

\subsection{Hiding Images in an Image}
\label{hiding-images-in-an-image}
\begin{figure}[t!]
    \begin{subfigure}{0.035\linewidth}
        \parbox[][][c]{\linewidth}{\centering\raisebox{0in}{\rotatebox{90}{GT}}}
    \end{subfigure}
    \begin{subfigure}{0.965\linewidth}
        \begin{subfigure}{0.166\linewidth}
			\includegraphics[width=0.98\linewidth]{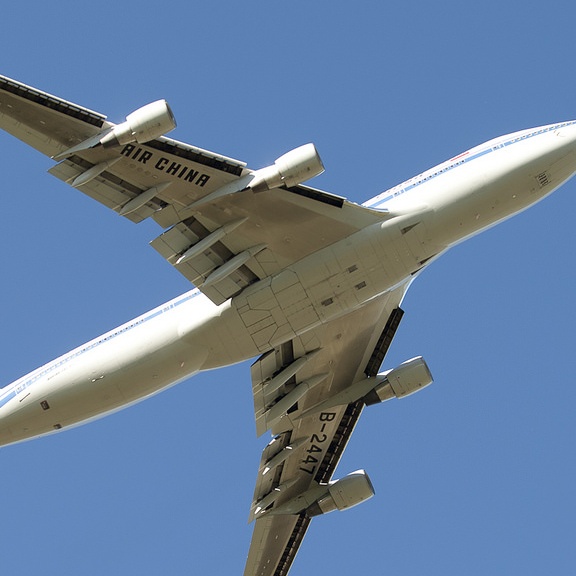}
    	\end{subfigure}%
        \begin{subfigure}{0.166\linewidth}
			\includegraphics[width=0.98\linewidth]{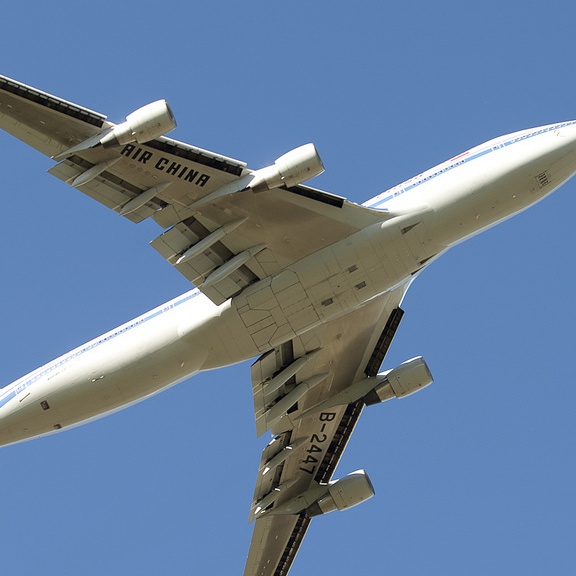}
    	\end{subfigure}%
        \begin{subfigure}{0.166\linewidth}
			\includegraphics[width=0.98\linewidth]{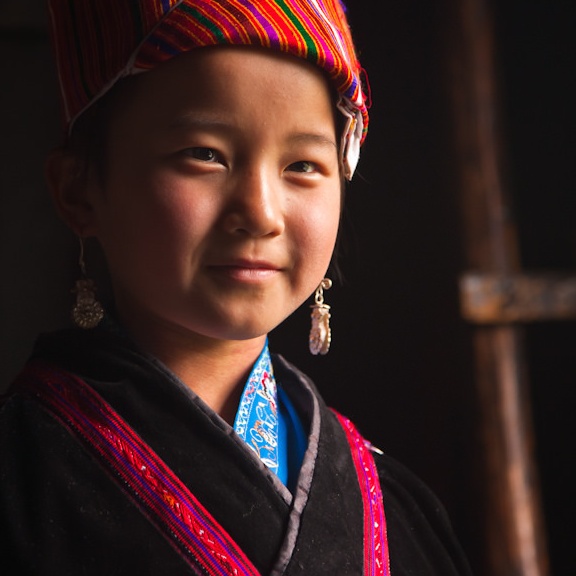}
    	\end{subfigure}%
        \begin{subfigure}{0.166\linewidth}
			\includegraphics[width=0.98\linewidth]{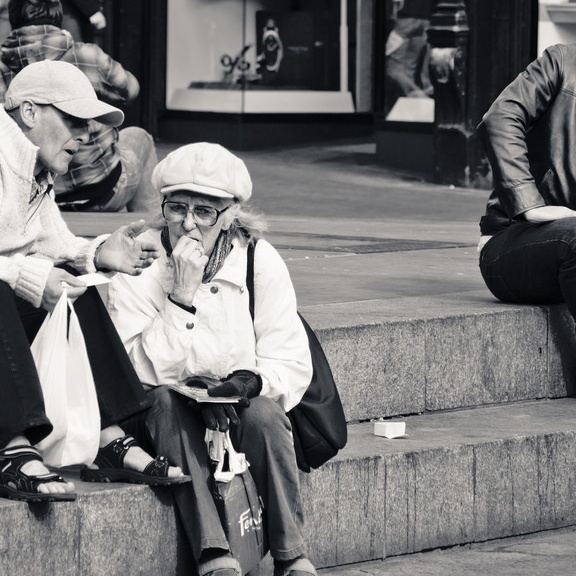}
    	\end{subfigure}%
        \begin{subfigure}{0.166\linewidth}
			\includegraphics[width=0.98\linewidth]{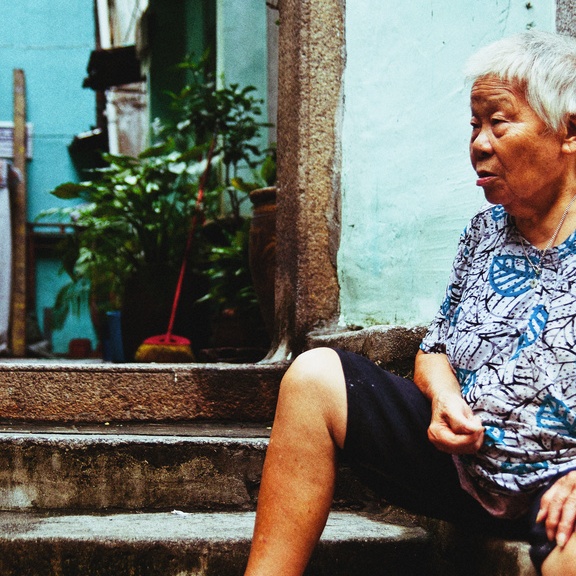}
    	\end{subfigure}%
	    \begin{subfigure}{0.166\linewidth}
			\includegraphics[width=0.98\linewidth]{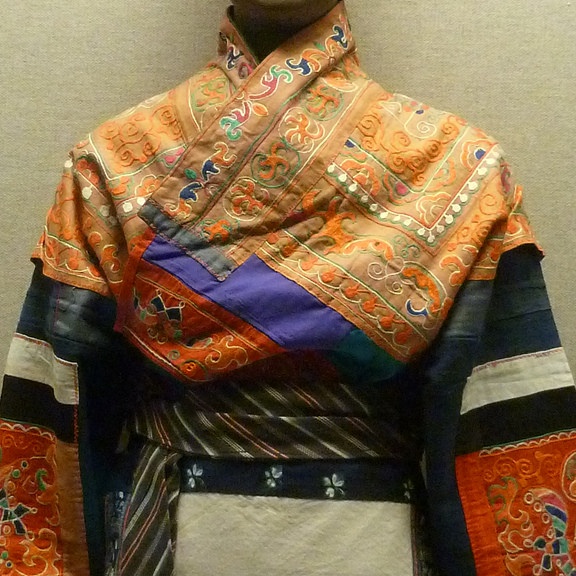}
    	\end{subfigure}%
	\end{subfigure}
	\\
    \begin{subfigure}{0.035\linewidth}
        \parbox[][][c]{\linewidth}{\centering\raisebox{0in}{\rotatebox{90}{Ours}}}
    \end{subfigure}
    \begin{subfigure}{0.965\linewidth}
        \begin{subfigure}{0.166\linewidth}
			\includegraphics[width=0.98\linewidth]{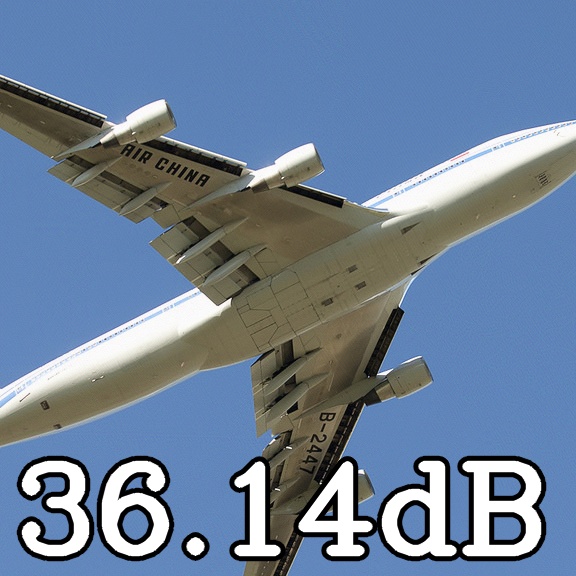}
    	\end{subfigure}%
        \begin{subfigure}{0.166\linewidth}
			\includegraphics[width=0.98\linewidth]{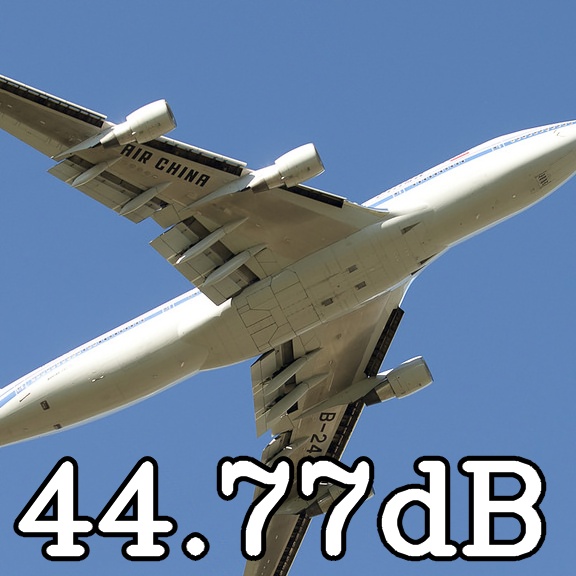}
    	\end{subfigure}%
        \begin{subfigure}{0.166\linewidth}
			\includegraphics[width=0.98\linewidth]{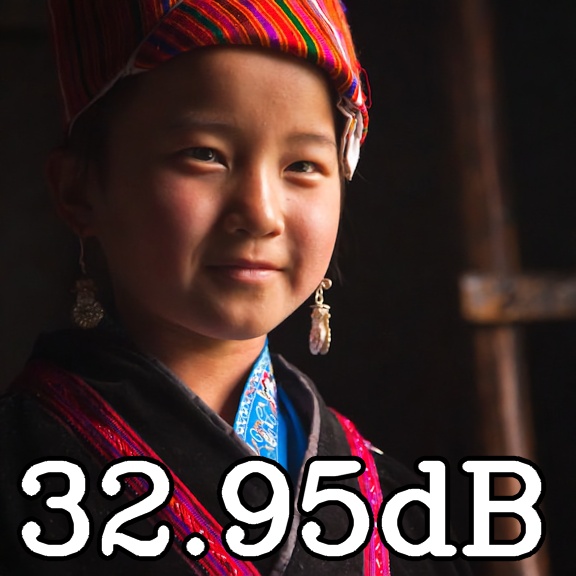}
    	\end{subfigure}%
        \begin{subfigure}{0.166\linewidth}
			\includegraphics[width=0.98\linewidth]{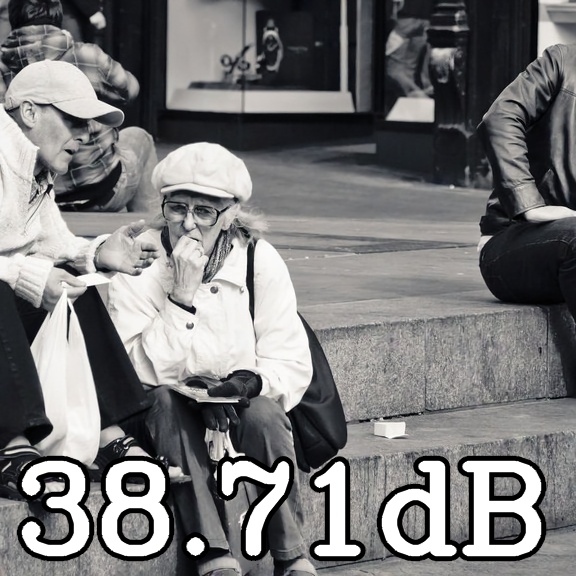}
    	\end{subfigure}%
        \begin{subfigure}{0.166\linewidth}
			\includegraphics[width=0.98\linewidth]{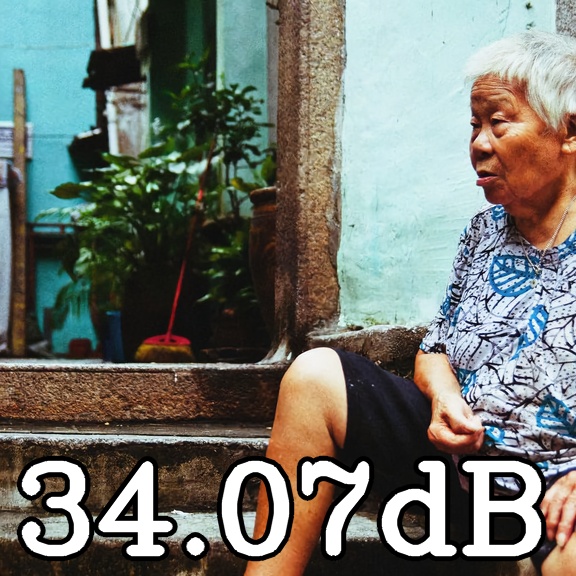}
    	\end{subfigure}%
	    \begin{subfigure}{0.166\linewidth}
			\includegraphics[width=0.98\linewidth]{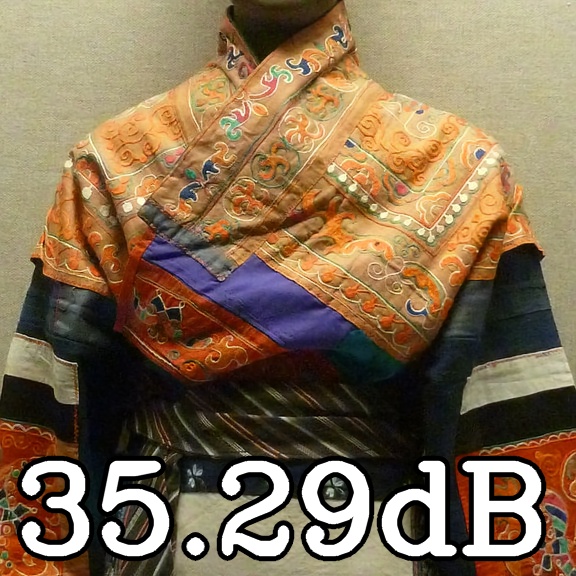}
    	\end{subfigure}%
	\end{subfigure}
	\\[2pt]
    \begin{subfigure}{0.035\linewidth}\hfil \end{subfigure}
    \begin{subfigure}{0.965\linewidth}
        \begin{subfigure}{0.155\linewidth}\hfil Embed.\end{subfigure}
        \begin{subfigure}{0.155\linewidth}\hfil Rev. 1\end{subfigure}
        \begin{subfigure}{0.155\linewidth}\hfil Rev. 2\end{subfigure}
        \begin{subfigure}{0.155\linewidth}\hfil Rev. 3\end{subfigure}
        \begin{subfigure}{0.155\linewidth}\hfil Rev. 4\end{subfigure}
        \begin{subfigure}{0.155\linewidth}\hfil Rev. 5\end{subfigure}
    \end{subfigure}
	\caption{A sample result of Hiding Images in Image}
	\label{fig:application3-visual}
\end{figure}

To show the generality of our proposed model, we try the hardest task to hide several unrelated images with our model, which can be viewed as a kind of stenography. We obtain general images from the Flicker 2W dataset~\cite{liu2020unified}. We conduct experiments to embed $2$, $3$, $4$, $5$ images into one image, and the numerical results are listed in Table~\ref{tab:application3-psnr-ssim}. From the results, we can see that our method achieves relatively good performance even when embedding $5$ images into one, demonstrating the strong generality of our method. From the visual results shown in Figure~\ref{fig:application3-visual}, despite the variety of colors and structures of the images, we can restore them with no viewable artifacts.

\section{Ablation Studies} 
\label{sec:ablation}
To ablate our network components and the applied frequency loss, we report some ablation results on three representative tasks in Table~\ref{tab:ablation}.
For AE, we use the network architecture proposed by Xia et al.~\cite{xia2018invertible} to represent general encoder-decoder based methods; for INNs, we adopt the network design and training strategy introduced by Xiao et al.~\cite{xiao2020invertible} to represent common INN based methods with auxiliary maps. We also present the results of our methods without relation module or frequency loss. For fair comparisons, all the models (unless otherwise specified) are trained with the applied frequency loss as discussed in Section~\ref{sec:method}, and we adjust the number of invertible blocks or CNN layers of different methods to have a similar number of parameters. 

The experiments show that our method outperforms general encoder-decoder style networks and common INNs with auxiliary maps. Intuitively, we know that there exists a trade-off relation between the embedding quality and the restoration quality. From the reported statistics, we can conclude that the frequency loss greatly contributes to the artifacts-free embedding for a significant quality boost with comparable restoration quality. Also, the proposed relation module works well to integrate with INNs to extract cross-image relationships and boost the performance.

\section{Conclusion and Discussion}
\label{sec:conclusion}
We present a generic framework IICNet for various reversible image conversion (RIC) tasks. IICNet maintains a task-independent and highly invertible architecture based on invertible neural networks (INNs), which can help greatly minimize the information loss during the conversion process. Due to strict invertibility, INNs have limitations in terms of nonlinear representation capacity and dimensional flexibility. The introduced relation module and the applied channel squeeze layer can greatly alleviate such limitations for better cross-image relation extraction and preserve the information-reserving ability of INNs.

IICNet yields state-of-the-art performance on some studied RIC tasks, such as spatial-temporal video embedding and mononizing binocular images. We also introduce and apply our IICNet on some unexplored tasks, which are embedding dual-view images and composition and decomposition. The success on the stenography task further shows the generalization of our IICNet. We hope the generalization and high performance of the proposed framework could help in more practical applications.

{\small
\bibliographystyle{ieee_fullname}
\bibliography{iicnet}
}

\end{document}